\documentclass[fleqn,10pt]{wlscirep}
\usepackage[utf8]{inputenc}
\usepackage[T1]{fontenc}
\usepackage{lineno}
\usepackage{multirow}
\usepackage{booktabs}
\usepackage{graphicx}
\usepackage{subfig}
\usepackage{subcaption}
\usepackage{array}
\usepackage{threeparttable}
\usepackage{float}

\title{COph100: A comprehensive fundus image registration dataset from infants \textcolor{black}{ constituting the "RIDIRP" database}}

\author[1,*]{Yan Hu}
\author[1]{Mingdao Gong}
\author[1]{Zhongxi Qiu}  
\author[1]{Jiabao Liu}
\author[1]{Hongli Shen}
\author[2]{Mingzhen Yuan}
\author[3]{Xiaoqing Zhang}
\author[1]{Heng Li}
\author[2]{Hai Lu}
\author[1,*]{Jiang Liu}
\affil[1]{Research Institute of Trustworthy Autonomous Systems and Department of Computer Science and Engineering, \\ Southern University of Science and Technology, Shenzhen, China}
\affil[2]{Beijing Tongren Eye Center, Beijing Tongren Hospital, Capital Medical University, Beijing Ophthalmology and Visual Sciences Key Laboratory, Beijing, China}
\affil[3]{Center for High Performance Computing and Shenzhen Key Laboratory of Intelligent Bioinformatics, Shenzhen Institute of Advanced Technology, Chinese Academy of Sciences,Shenzhen, China}

\affil[*]{Corresponding author(s): Yan Hu and Jiang Liu (huy3@sustech.edu.cn, liuj@sustech.edu.cn)}

\begin{abstract}
Retinal image registration is vital for diagnostic therapeutic applications within the field of ophthalmology. Existing public datasets, focusing on adult retinal pathologies with high-quality images, have limited number of image pairs and neglect clinical challenges. To address this gap, we introduce COph100, a novel and challenging dataset known as the Comprehensive Ophthalmology Retinal Image Registration dataset for infants with a wide range of image quality issues \textcolor{black}{constituting the public "RIDIRP" database \cite{timkovivc2024retinal}. COph100 consists of 100 eyes, each with 2 to 9 examination sessions, amounting to a total of 491 image pairs carefully selected from the publicly available dataset}. We manually labeled the corresponding ground truth image points and provided automatic vessel segmentation masks for each image. We have assessed COph100 in terms of image quality and registration outcomes using state-of-the-art algorithms. This resource enables a robust comparison of retinal registration methodologies and aids in the analysis of disease progression in infants, thereby deepening our understanding of pediatric ophthalmic conditions.

\end{abstract}
\begin{document}

\flushbottom
\maketitle

\thispagestyle{empty}


\section*{Background \& Summary}
\noindent Retinal image registration plays a pivotal role in disease diagnosis \cite{avants2008symmetric}, image-guided surgery \cite{alam2018medical}, monitoring of disease progression \cite{javaid2016visual}, and image fusion \cite{james2014medical}. The primary objective of image registration is to spatially align a query (source) image with a reference (target) image by determining a geometric transformation that accurately matches the corresponding features or structures between the two images. In the realm of ophthalmology, retinal image registration has become an indispensable tool, facilitating precise tracking of temporal changes in retinal anatomy and the alignment of different imaging modalities, which is crucial for evaluating disease evolution and treatment outcomes \cite{nie2024medical,pan2021retinal}. 

In the field of ophthalmology, various imaging modalities are used, including color fundus (CF), optical coherence tomography (OCT), fluorescein angiography (FA). The integration of these diverse imaging modalities via registration enhances the comprehensive analysis of retinal pathologies. Numerous studies have been dedicates to proposing retinal image registration algorithms, thereby propelling the advancement of this domain. As delineated in several review papers \cite{nie2024medical,pan2021retinal}, these algorithms can be categorized into multi-modal and single modal image registration based on the input image modality. The corresponding datasets are summarized in Table \ref{tab:datasetreview}, all of which come with the ground truth data for registration purposes. Datasets such as RODREP \cite{adal2015accuracy} (\href{http://www.rodrep.com/data-sets.html}{http://www.rodrep.com/data-sets.html}), TeleOphta\cite{decenciere2013teleophta}(\href{https://www.adcis.net/en/third-party/e-ophtha/}{https://www.adcis.net/en/third-party/e-ophtha/}), and VARIA \cite{ortega2009retinal} (\href{http://www.varpa.es/research/biometrics.html}{http://www.varpa.es/research/biometrics.html}), which were originally created for application other \textcolor{black}{than} registration, are excluded from the table.

Recent advancements in retinal imaging have led researchers to propose various multi-modal image registration datasets. For instance, Lee et al. \cite{lee2015registration} introduced a private dataset in 2015, which includes CF and OCT images. In 2020, Almsasi et al.\cite{almasi2020registration} presented a FA-scanning laser ophthalmoscopy (FA-SLO) dataset, from patients with diabetic retinopathy. The PRIME-FP20 dataset \cite{ding2020weakly}(\href{https://ieee-dataport.org/open-access/prime-fp20-ultra-widefield-fundus-photography-vessel-segmentation-dataset}{https://ieee-dataport.org/open-access/prime-fp20-ultra-widefield-fundus-photography-vessel-segmentation-dataset}), introduced in 2021, comprises ultra-wide-field fluorescence angiography (FA) and fundus photography (FP) images. FOCTAIR \cite{martinez2021robust}(\href{http://www.varpa.es/research/ophtalmology.html}{http://www.varpa.es/research/ophtalmology.html}) contains FA and optical coherence tomography angiography (OCTA) images from retinal vein occlusion (RVO) patients. In 2024, MEMO \cite{Wang:24}(\href{https://chiaoyiwang0424.github.io/MEMO/}{https://chiaoyiwang0424.github.io/MEMO/}) was introduced, which consists of \textcolor{black}{Erythrocyte-mediated Angiography}(EMA) and OCTA images from normal patients.

While additional imaging modalities such as OCT and FA are employed as necessary for diagnostic or therapeutic purposes, color fundus photography is still the most frequently utilized data type in clinical settings. For example, FLoRI21 \cite{ding2021flori21} (\href{https://ieee-dataport.org/open-access/flori21-fluorescein-angiography-longitudinal-retinal-image-registration-dataset}{https://ieee-dataport.org/open-access/flori21-fluorescein-angiography-longitudinal-retinal-image-registration-dataset})includes ultra-wide-field fundus photography from 15 patients, presenting a significant challenge for registration due to its extensive field of view. The LSFG dataset \cite{sivaraman2024retinaregnet}, from 2024, encompasses images from 15 patients with uveal melanoma. Given that color fundus photography remains the most commonly used imaging modality in hospitals, the FIRE dataset \cite{hernandez2017fire}(\href{https://projects.ics.forth.gr/cvrl/fire/}{https://projects.ics.forth.gr/cvrl/fire/}) has been predominantly evaluated. It is categorized into three classes based on the extent of image overlap and the presence or absence of anatomical features. The majority of image pairs within the FIRE dataset are utilized for super-resolution and mosaicking purposes. Only 14 image pairs were acquired from different examinations of retinopathy, yet they exhibit significant overlap. All images in the FIRE dataset are of high resolution. With the integration of deep learning algorithms into fundus image registration, the accuracy of the dataset has already achieved a high level, with a reported 0.85 mean Area Under the Curve (mAUC) and RMSE less than 2.9 \cite{wang2024superjunction}. Consequently, in response to current demands, \textcolor{black}{we construct a challenging monomodal fundus image registration dataset COph100, which contains 100 eyes and 491 pairs of images from Retinopathy of Prematurity (ROP) infants. Given the current publicly available ROP datasets \cite{timkovivc2024retinal}(https://doi.org/10.6084/m9.figshare.c.6626162.v1), this paper primarily discusses how to utilize these existing public datasets to further pursue research of one’s interest.}

\begin{table}[]
    \centering
    \caption{The dataset and dataset review.}
    \begin{tabular}{|c|c|c|c|c|c|c|c}
\hline        Dataset & Image Modality &Num of pairs(Num of eyes) &Diseases & Availability & Publish time\\
\hline Lee \cite{lee2015registration} & CF-OCT & 52(52)  & - &Private &2015\\
\hline Almasi \cite{almasi2020registration} &FA-SLO &21(36)  &Diabetic Retinopathy &Private &2020 \\
\hline PRIME-FP20 \cite{ding2020weakly} & UWF FA-FP &15(15)& - & Free &2021\\
\hline FOCTAIR \cite{martinez2021robust} & FA-OCTA & 172(29) & RVO &On demand &2021\\
\hline MEMO\cite{Wang:24} & EMA-OCTA &30(30)  & No & Free & 2024 \\
\hline FIRE\cite{hernandez2017fire} & CF & 134 (39)  & Diabetic related& Free &2017\\
\hline FLoRI21 \cite{ding2021flori21} & UWF FA &15(15) &- & Free &2021 \\
\hline LSFG\cite{sivaraman2024retinaregnet} &LSFG &15(15) & Uveal Melanoma & Private &2024\\
        \hline 
        \textcolor{black}{\textbf{COph100 (Ours)}} & \textcolor{black}{RetCam} &\textcolor{black}{491(100)} & \textcolor{black}{Retinopathy of Prematurity} &\textcolor{black}{Public} &\textcolor{black}{2024}\\
\hline
    \end{tabular}
    \begin{tablenotes}
        \footnotesize
        \item[*] CF: Color Fundus; OCT/A: Optical Coherence Tomography/Angiography; UWF: Ultra-Wide Field;  FA: Fluorescein Angiography; FP: Fundus Photography; SLO: Scanning Laser Ophthalmoscopy; EMA: Erythrocyte-mediated Angiography; LSFG: Laser Speckle Flowgraphy. 
    \end{tablenotes}
    \label{tab:datasetreview}
\end{table}

To the best of our knowledge, the COph100 dataset is the first retinal registration dataset specifically focused on disease progression in infants. Its potential impact, which addresses several existing limitations, can be outlined as follows: First, \textit{Minimal apprarance variablity:} A common limitation of most retinal image registration datasets is the lack of significant appearance variation, which does not adequately reflect the diversity encountered in clinical practice. The COph100 dataset, however, includes image pairs with substantial appearance differences resulting from variations in acquisition time, patient condition, and imaging environments. This diversity enhances the generalizability of registration algorithms, enabling them to perform more robustly in real-world clinical settings where image characteristics vary widely. Second, \textit{Pediatric pathologies focus:} Existing datasets primarily focus on adult retinal pathologies, thereby neglecting the distinct features and challenges of pediatric retinal diseases. The COph100 dataset addresses this gap by concentrating on pediatric retinal pathologies, expanding the clinical relevance and applicability of registration algorithms beyond the adult population. Third, \textit{Scalability and population diversity:} Current retinal image registration datasets are limited in scale, often involving fewer than 52 eyes, which can introduce bias in algorithm performance. In contrast, the COph100 dataset offers greater scalability and includes a more diverse patient population. This increased diversity provides a more robust basis for developing algorithms that are generalizable across different demographic groups, leading to improved clinical outcomes.
Therefore, our proposed COph100 dataset represents a significant advancement in retinal image registration by providing a more diverse, pediatric-focused, and scalable dataset that improves the robustness and clinical applicability of registration algorithms \textcolor{black}{based on the publicly available dataset\cite{timkovivc2024retinal}}.

\section*{Methods}

\subsection*{Data Preparation} 
\subsubsection*{Image datasets}
In this study, we aim to utilize publicly available retinal datasets to construct a comprehensive retinal image registration dataset. Our first step involves identifying datasets that include image pairs, varying poses or follow-up examinations, such as RODREP \cite{adal2015accuracy}, TeleOphta\cite{decenciere2013teleophta}, and a dataset proposed by \textcolor{black}{Timkovic et al.}\cite{timkovivc2024retinal}(\href{https://doi.org/10.6084/m9.figshare.c.6626162.v1}{https://doi.org/10.6084/m9.figshare.c.6626162.v1}). Certain datasets, like the 2021 \textcolor{black}{Retinopathy of Prematurity (ROP)} dataset \cite{agrawal2021assistive}, which only contains images from different \textcolor{black}{poses}, are not considered for this study.
We compare these dataset with FIRE, as shown in Table \ref{tab:datasetpre}. The RODREP dataset, proposed in 2015, consists of fundus images from 70 diabetic patients with two examinations. TeleOptha, introduced in 2013, includes images from one or both eyes of healthy and diabetic retinopathy patients, with one or two examinations. The dataset proposed by \textcolor{black}{Timkovic et al.\cite{timkovivc2024retinal}} in 2024 contains 2-9 examinations of premature infants with ROP. It includes images captured at three different resolutions by various devices, providing a diverse set of eyes. The image resolution for RODREP, TeleOphtha and two sections ($1240 \times 1240$ and $1440 \times 1080$) of the \textcolor{black}{Timkovic et al.} dataset\cite{timkovivc2024retinal} is relatively high. 

Several factors were considered in selecting a suitable dataset for our challenging COph100 dataset. First, the dataset should include a large number of patient eyes. Second, it should contain images from patients with at least two examinations. Finally, the image resolution should not be excessively high, as high-quality retinal images are not always obtainable in clinical practice. Based on these criteria, we have decided to construct the challenging COph100 dataset using the lower-resolution images ($640 \times 480$) from the dataset proposed by \textcolor{black}{Timkovic et al.\cite{timkovivc2024retinal}}
\begin{table}[!h]
        \caption{Preparation of datasets. \textcolor{blue}{FIRE is a commonly used image registration dataset. The datasets of RODREP, TeleOphta, Timkovic et al. contains 2 or more examinations, which align with the purpose of our paper - to construct a registration dataset that contains data from different patient examination periods.}}
        \centering
        \begin{threeparttable}
            \begin{tabular}{|c|c|c|c|c|c|c|}
                \hline Dataset& FIRE  & RODREP &TeleOphta &  \multicolumn{3}{c|}{\textcolor{black}{Timkovic et al.}} \\
                \hline Example & \begin{minipage}{.12\textwidth}
                \includegraphics[width=20mm,height=10mm]{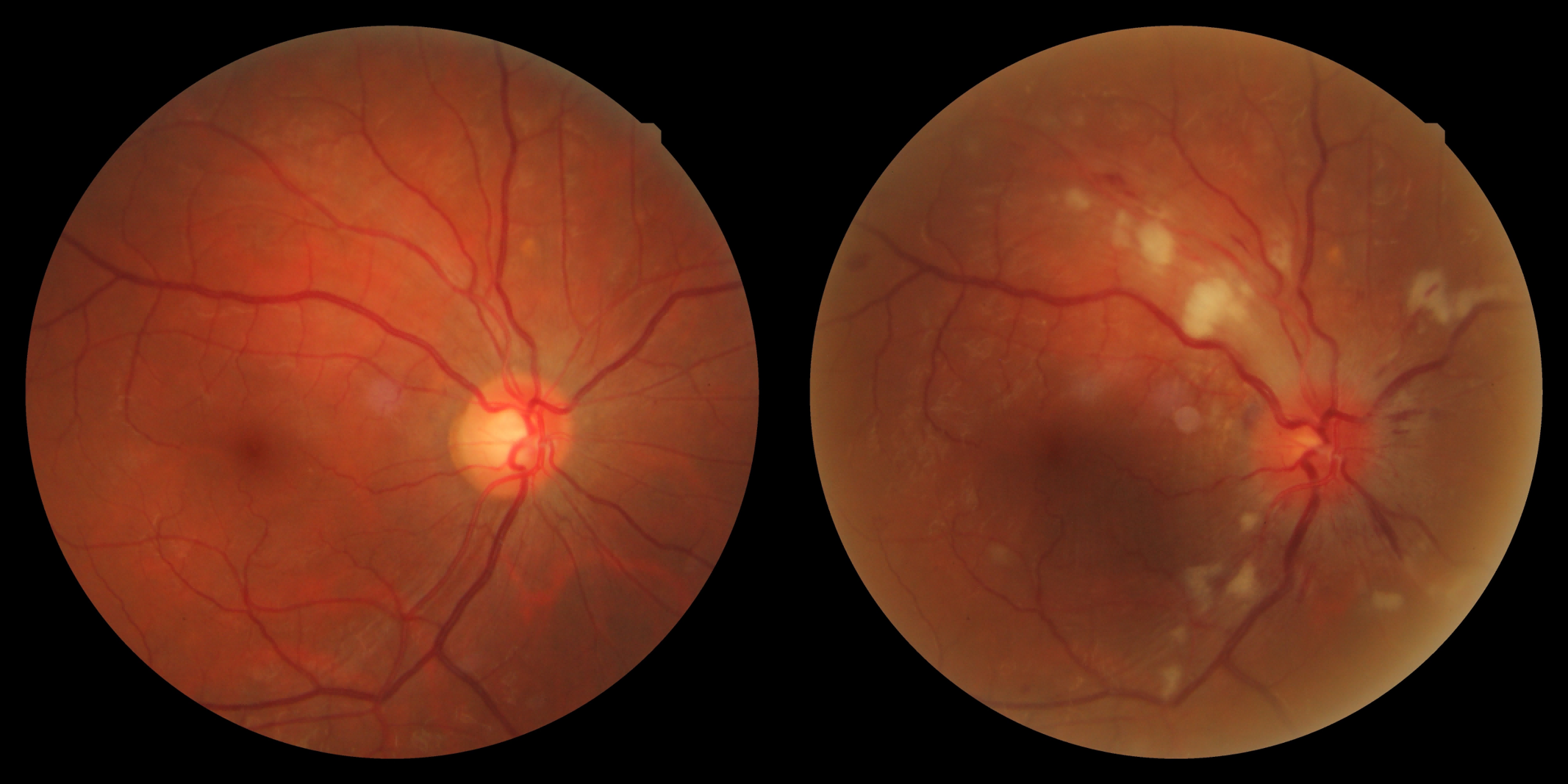}   
                \end{minipage}  & \begin{minipage}{.12\textwidth}
                \includegraphics[width=20mm,height=10mm]{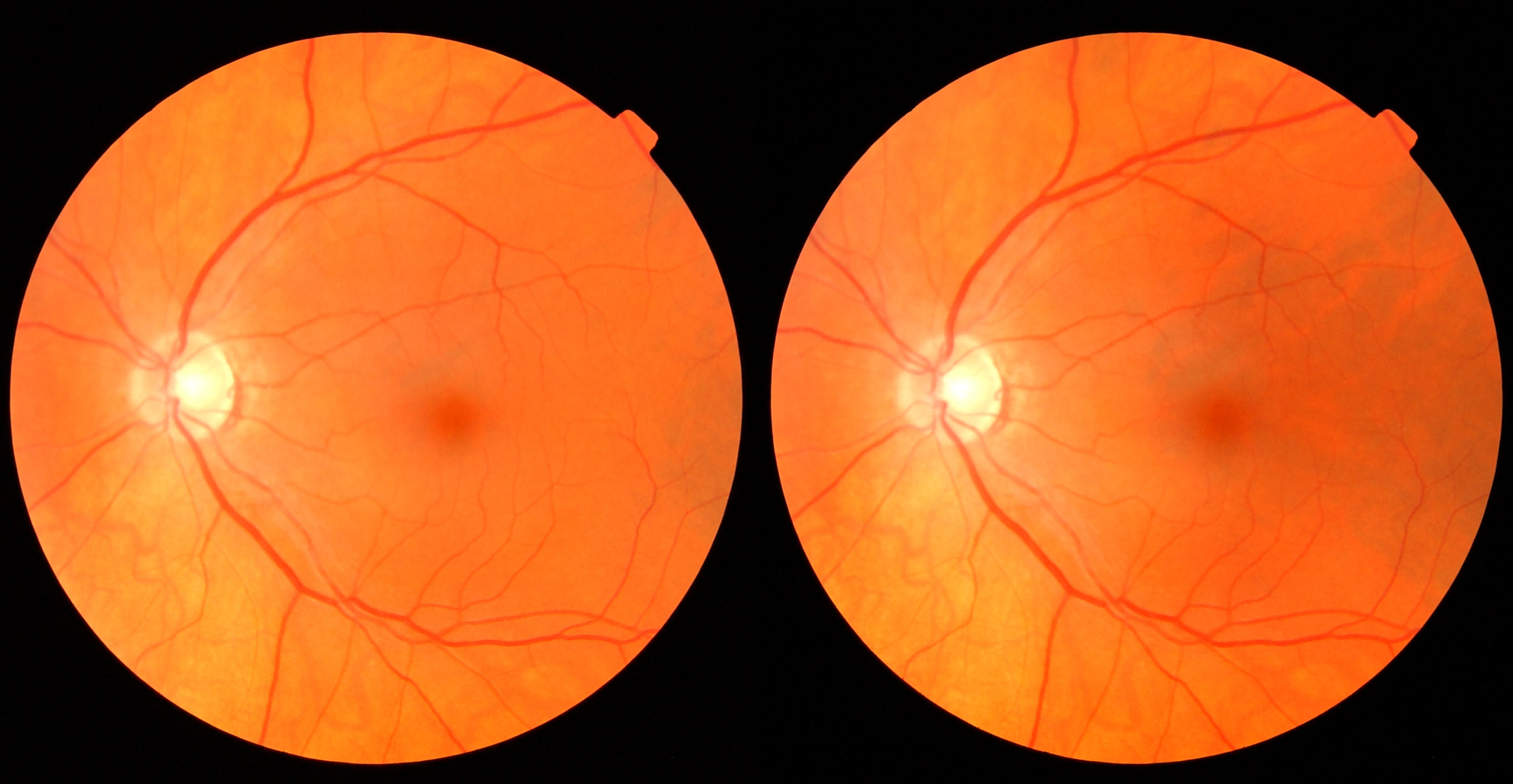} 
                \end{minipage}  &\begin{minipage}{.12\textwidth}
                \includegraphics[width=20mm,height=10mm]{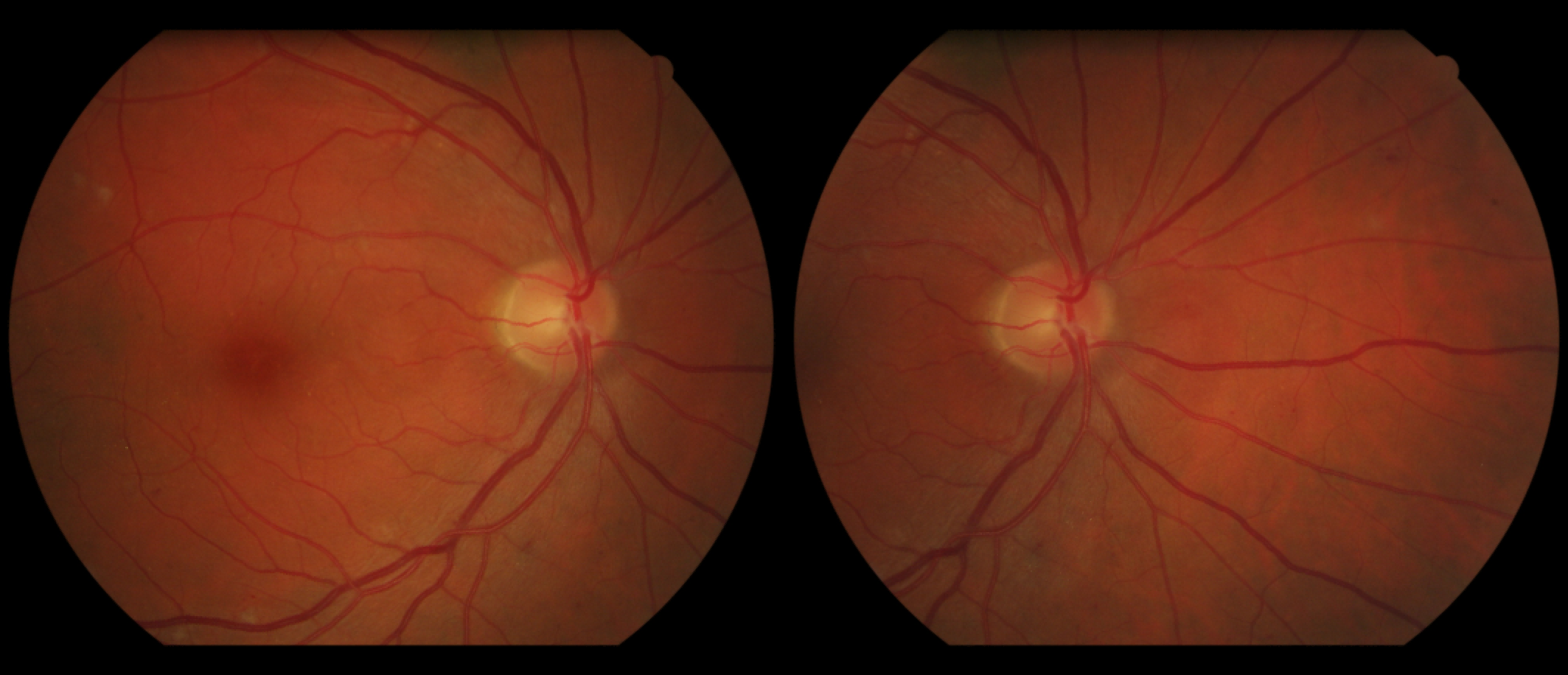}            
                \end{minipage}  
                & \multicolumn{3}{c|}{\begin{minipage}{.12\textwidth}
                \includegraphics[width=20mm,height=10mm]{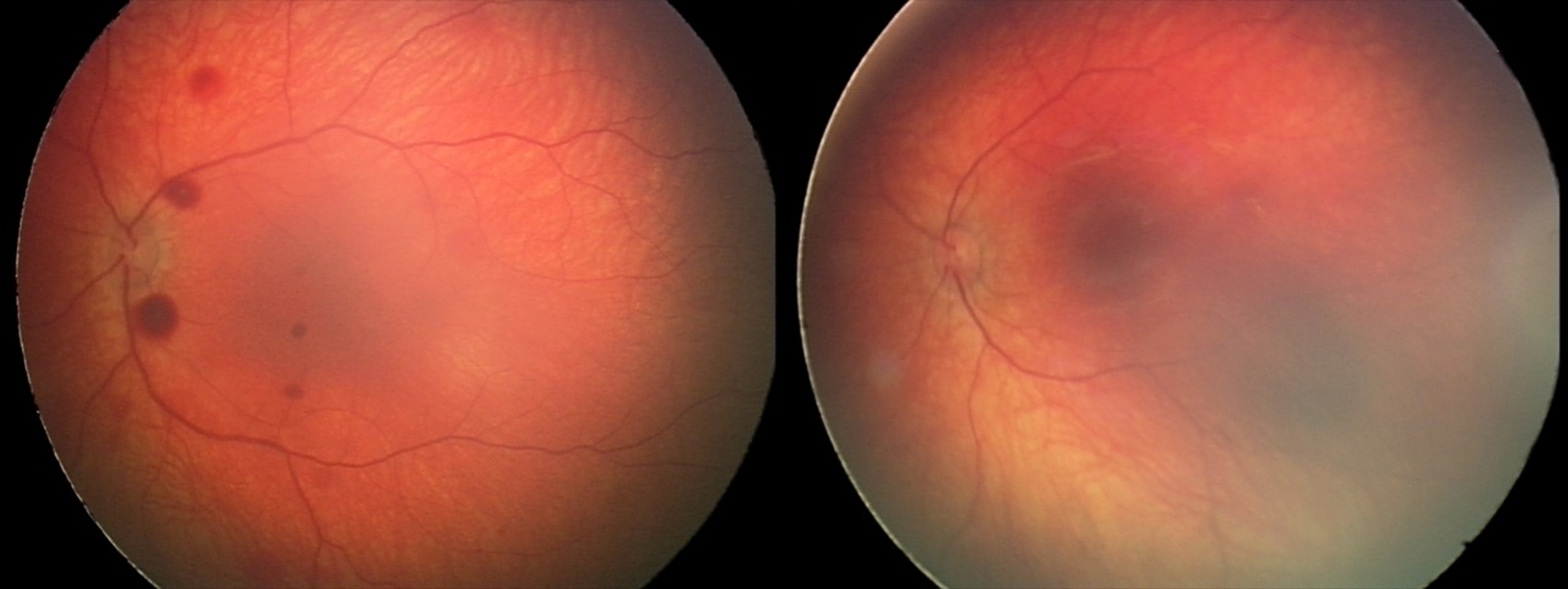}   
                \end{minipage}  
                \begin{minipage}{.12\textwidth}
                \includegraphics[width=20mm,height=10mm]{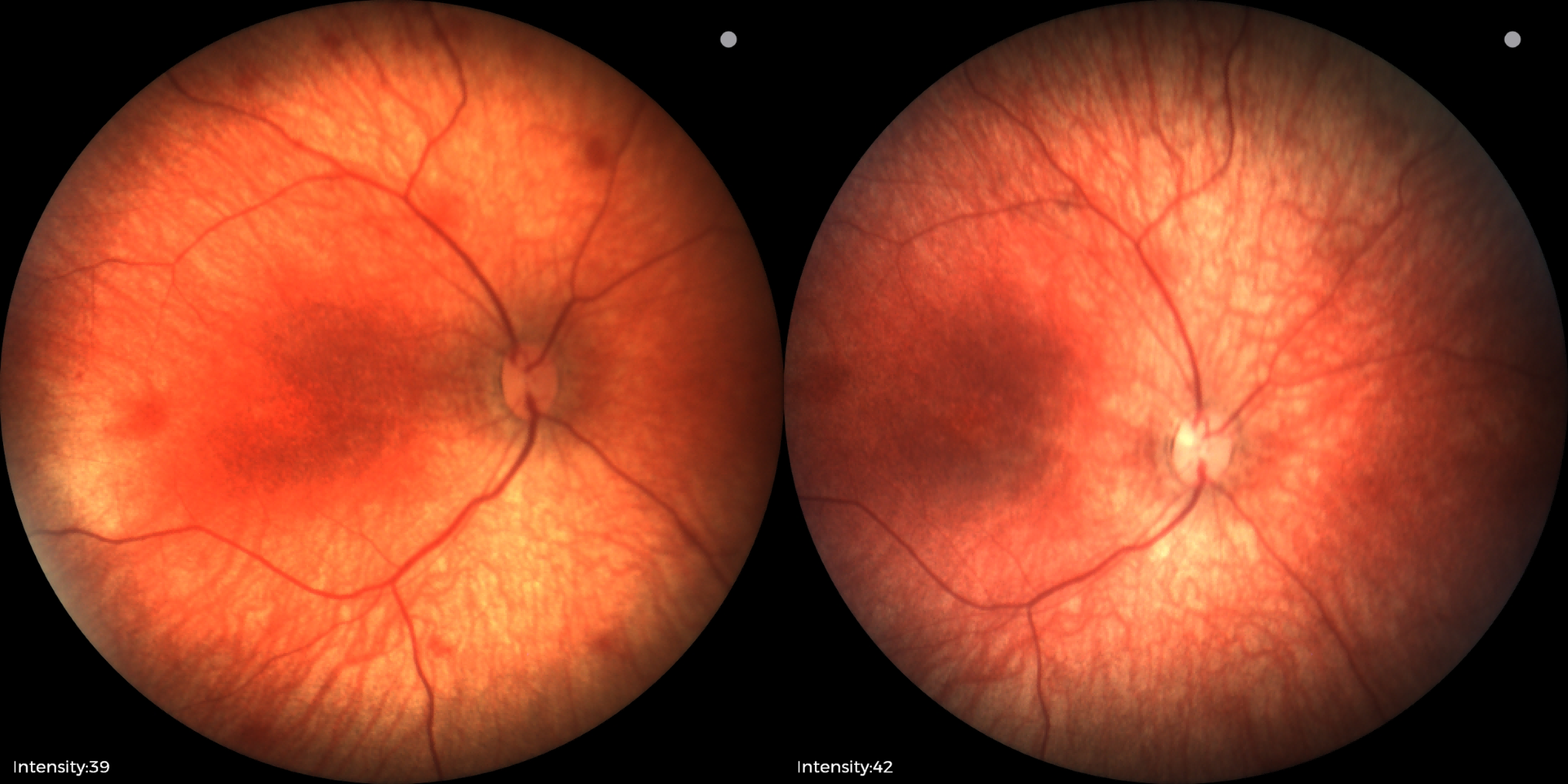}   
                \end{minipage} 
                \begin{minipage}{.12\textwidth}
                \includegraphics[width=20mm,height=10mm]{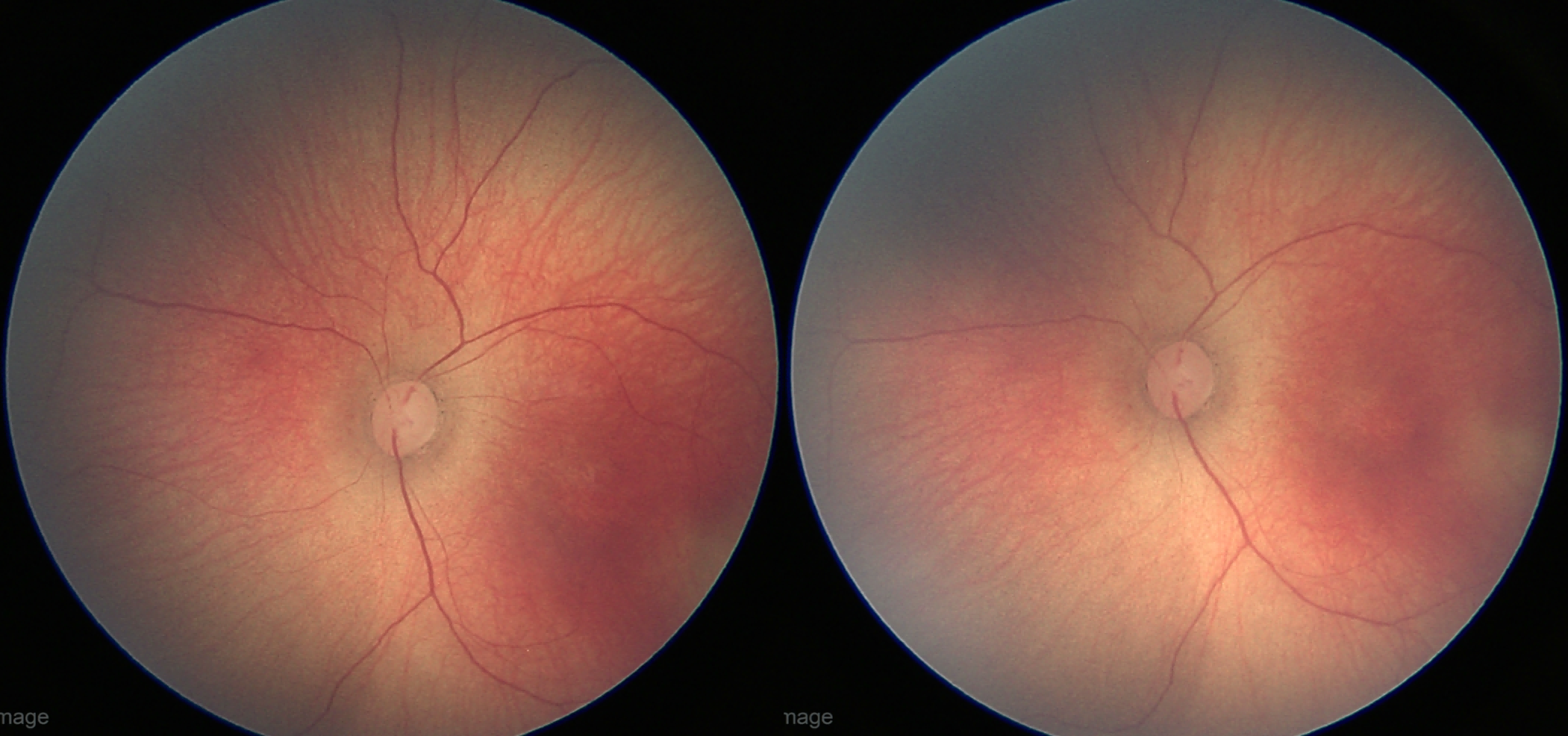}   
                \end{minipage}} \\
                \hline Resolution & 2912*2912  &2000*1312 &2544*1696   & \multicolumn{3}{c|}{640*480,1240*1240,1440*1080}\\
                \hline Eye Num* &  14  &140 & 119 &\multicolumn{3}{c|}{118,86,28} \\
                \hline Examinations &2 & 2 &2 &\multicolumn{3}{c|}{ 2-9} \\
                \hline
            \end{tabular}
    \begin{tablenotes}
        \footnotesize
        \item[*]  We just list the number of eyes with 2 or above examinations in datasets. 
    \end{tablenotes}
    \end{threeparttable}
    \label{tab:datasetpre}
\end{table}

\begin{figure}[h]
    \centering
    \includegraphics[width=0.75\textwidth]{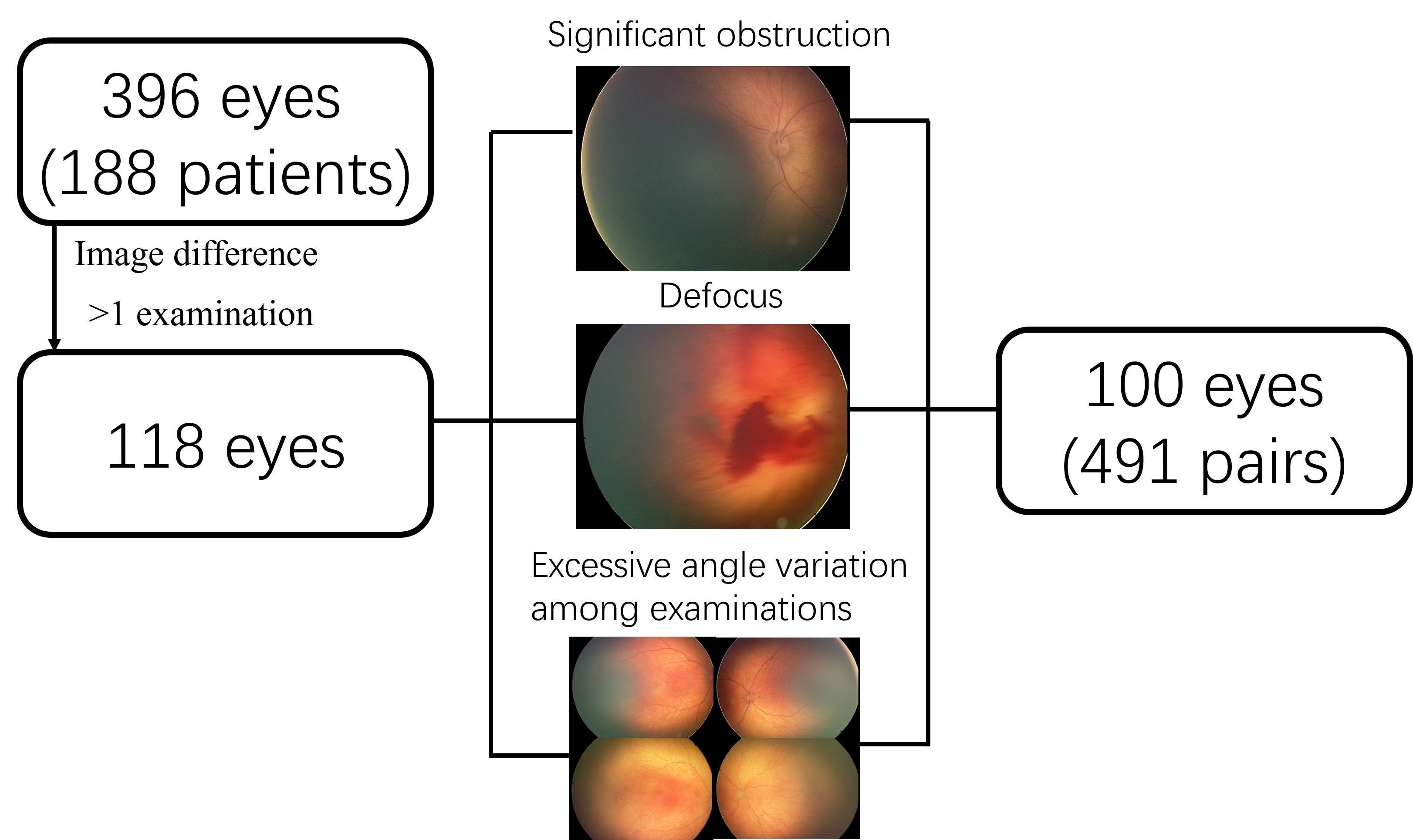}
    \caption{Summary diagram of inclusion-exclusion criteria.}
    \label{fig:inexclusion}
\end{figure}

\subsubsection*{Inclusion-exclusion criteria}
Our study primarily focuses on image registration across multiple examinations, aiming to analyze disease progression over time. The specific selection criteria are outlined in Fig. \ref{fig:inexclusion}. The original ROP grading dataset includes 396 eyes from 188 patients. To construct a more challenging registration dataset than FIRE, we deliberately selected images with greater appearance differences, such as variations in image resolution, illumination, compared to those in the FIRE dataset. Additionally, to assess disease progression effectively, we excluded patients with only a single examination, ultimately narrowing the dataset down to 118 eyes.

Capturing retinal images from infants, unlike adults, poses significant challenges. Infants cannot follow instructions such as maintaining fixation on a point or keeping their eyes still, which complicates the process of obtaining high-quality fundus images. The resulting images often suffer from large occlusions caused by diseases or eyelids, poor focus, or significant rotation and translation between examinations, leading to limited image overlap. Furthermore, since physicians need to monitor changes in lesion areas over time during diagnosis, we made efforts to select image pairs that consistently include the lesion area at different time points. This selection process is intended to facilitate future analysis of disease progression. Ultimately, we obtained 100 eyes with 491 image pairs for our COph100 dataset.


\begin{figure}[h]
    \centering
    \includegraphics[width=0.85\textwidth]{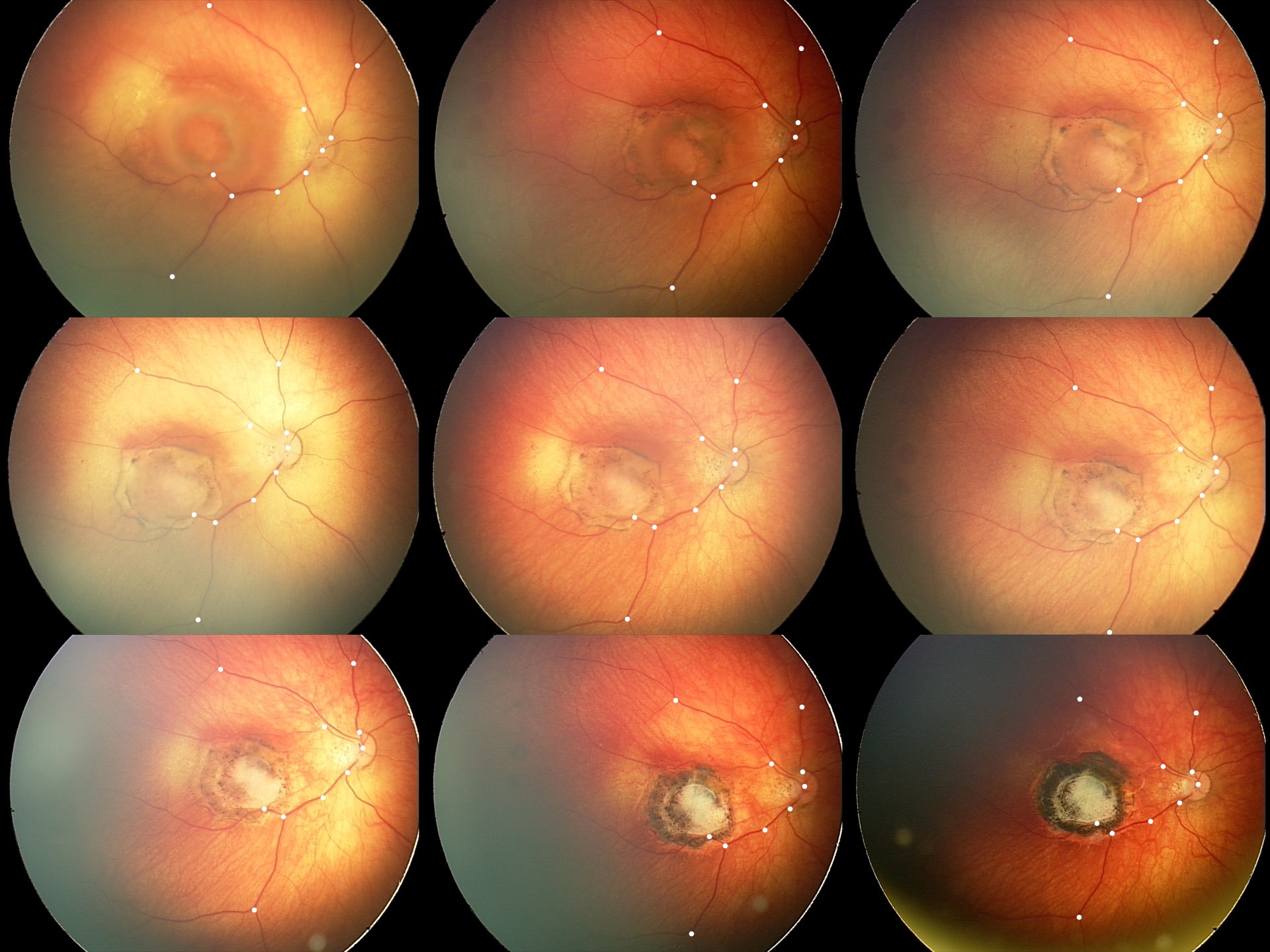}
    \caption{Image from patient with 9 examinations. The challenges include blur, obstruction, large lesions, illumination and color changes.}
    \label{fig:pointGT}
\end{figure}

\begin{figure}[h]
    \centering
    \includegraphics[width=0.8\textwidth]{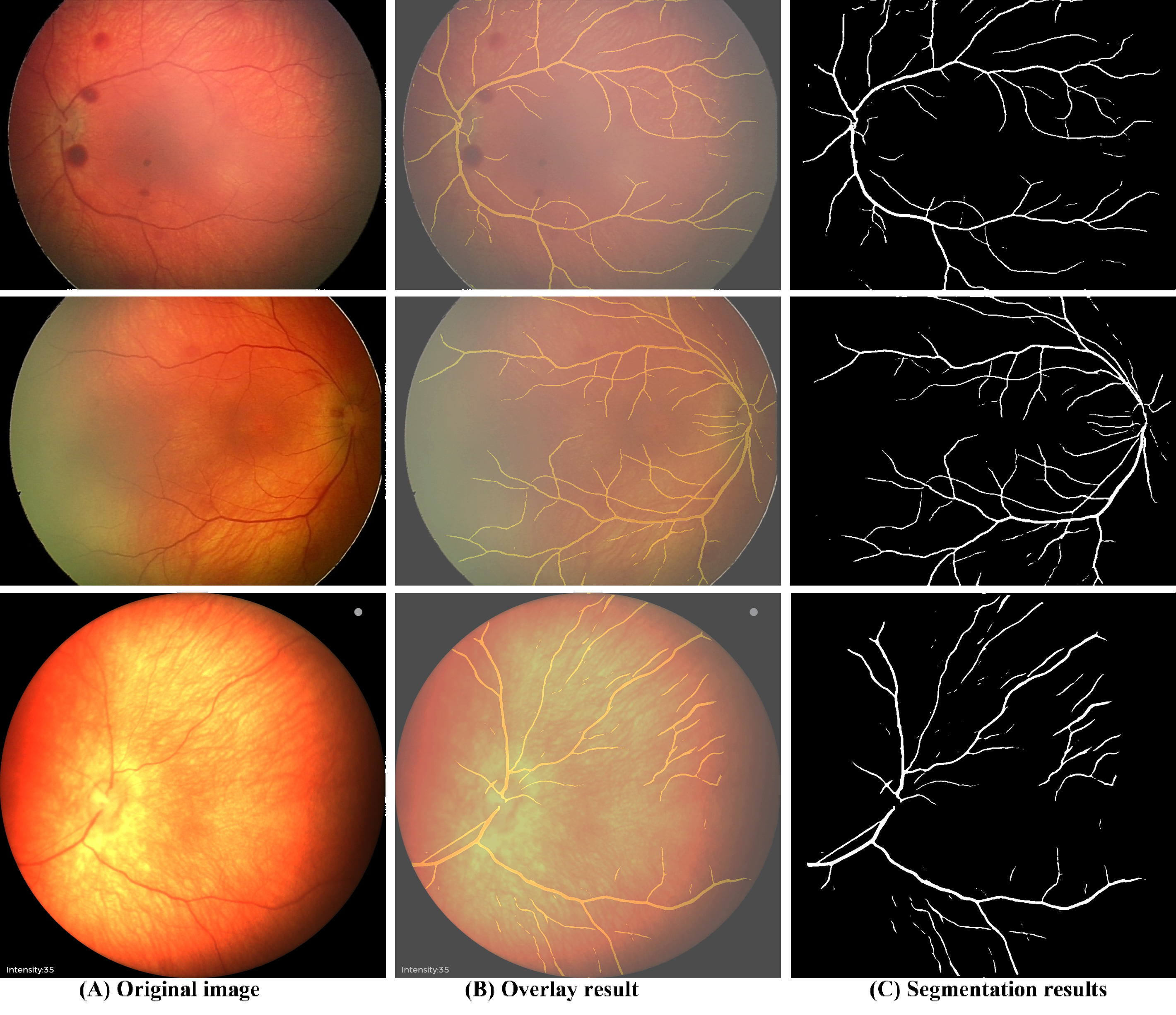}
    \caption{Visualization samples of Retinal vessel segmentation results.}
    \label{fig:vessel}
\end{figure}

\begin{table}[h]
        \caption{The distribution of examination sessions within our COph100 dataset.}
        \begin{center}
     \begin{tabular}{|c|c|p{13cm}|}
\hline
\textbf{Examinations} & \textbf{Eye Num} & \textbf{Folder Name} \\
\hline
\multirow{2}{*}{2} & \multirow{2}{*}{33} & '03', '03-1', '04', '09', '09-1', '12', '12-1', '14', '14-1', '16', '17','17-1', '20-1', '22', '22-1', \\
& &'23', '24', '24-1', '25', '25-1', '36', '36-1', '45', '45-1',  '54', '54-1', '62', '62-1', '63', '67',\\
& & '67-1', '68', '68-1' \\
\hline
\multirow{2}{*}{3} & \multirow{2}{*}{43} & '01', '02', '02-1', '05', '05-1', '07', '11', '11-1', '13', '13-1', '19', '20', '21', '21-1', '28', \\ & &'28-1', '32', '44', '44-1', '47', '47-1', '48', '48-1', '49',  '49-1', '50', '50-1', '56', '56-1', '57', '57-1', '58', '58-1', '59', '59-1', '61', '61-1', '64', '64-1', '65', '65-1', '76', '76-1' \\
\hline
4 & 14 & '08', '08-1', '27', '40', '41', '51', '51-1', '66-1', '69', '69-1', '70', '70-1', '71', '71-1' \\
\hline
5 & 3 & '10', '26', '26-1' \\
\hline
6 & 1 & '66' \\
\hline
8 & 2 & '52', '52-1' \\
\hline
9 & 4 & '31', '31-1', '55', '55-1' \\
\hline
\end{tabular}
        \end{center}   
    \label{tab:examinationlist}
\end{table}

\subsection*{Data processing}
\subsubsection*{Registration Groundtruth}
In image registration research, the evaluation of methods is typically conducted using both quantitative and qualitative metrics. Quantitative assessments often rely on either manually designated or automatically detected control points, which enable a concise representation of registration error as a single numerical value. Other quantitative methods, such as those that compute a transformation matrix, express error in terms of multiple parameters. On the other hand, qualitative evaluations focus primarily on the visual inspection of vessel alignment within the images, which necessitates expert input and lacks the ability to provide objective, quantitative comparisons across different registration techniques.

In the proposed COph100 dataset, we offer the actual corresponding points needed to compute registration errors, which are termed as control points, as illustrated in Fig. \ref{fig:pointGT}. For each eye, the location of a control point $j$ in the image of first examination is denoted as $E_{1j}$, and the corresponding point in the image of other examination $i$ as $E_{ij}$. The registration process uses the point $E_{ij}$ as input and transforms it to the new set of coordinates referred to as $M_{ij}$. Thus, the $M_{ij}$ coordinates represent the post-registration positions of the $E_{ij}$ points. Ideal registration results in the points $M_{ij}$ and $E_{ij}$ being indistinguishable, with a pixel distance of 0 between them. In Fig. \ref{fig:pointGT}, each point corresponds to each other in all 9 examinations. The ground truth points are primarily marked around the vessel intersections.

\subsubsection*{Retinal Vessel Segmentation} The vessel structure plays a crucial role in analyzing the condition of the retina. To extract the vessel structure, we developed a vessel segmentation model based on SS-MAF\cite{10274145}(\href{https://github.com/Qsingle/imed\_vision}{https://github.com/Qsingle/imed\_vision}). Following prior research \cite{owler2021influence}, we selected the green channel from the RGB image format as the input for the model, and we trained the model using the publicly available FIVES dataset\cite{jin2022fives}(\href{https://doi.org/10.6084/m9.figshare.19688169.v1}{https://doi.org/10.6084/m9.figshare.19688169.v1}). The training settings were consistent with \cite{10274145}, where the 600 images in the training dataset were split with a 4:1 ratio for training and validation purposes. Upon training, the model achieved performance metrics of 91.56\% in Dice score, 89.81\% in sensitivity (SE), and 89.32\% in bookmaker informedness (BM) at the test set of the FIVES dataset. The segmentation results are visualized in Fig.\ref{fig:vessel}, demonstrating that the model successfully captures nearly all of the vessel structures. The collaborating opthalmologists also evaluated the segmentation results and concluded that they are sufficient for disease progression analysis. Since the original grading dataset proposed by Timkovic et al.\cite{timkovivc2024retinal} includes a section on lesion segmentation, we did not extract the lesion areas separately in our study.

Formulations for the evaluation metrics are as follows:
\begin{align}
    &Dice = \frac{2 \times TP}{2 \times TP+FN+FP}, SE = \frac{TP}{TP+FN}, BM = \frac{TP}{TP+FN}+\frac{TN}{TN+ FP}-1 
\end{align}
where $TP$, $TN$, $FP$, and $FN$ are the true positive, true negative, false positive, and false negative respectively.

\begin{figure}[!h]
    \centering
    \includegraphics[width=0.8\textwidth]{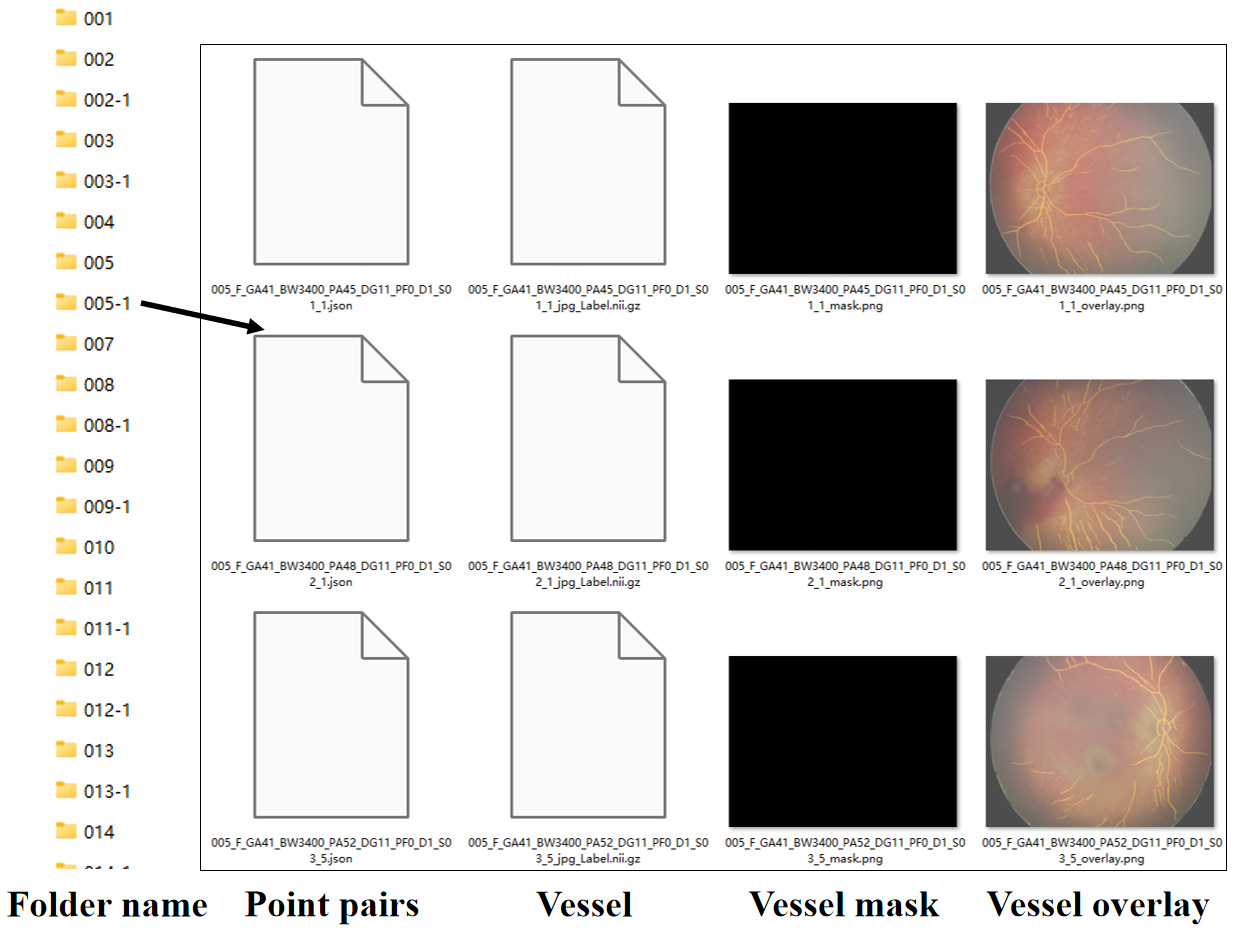}
    \caption{The details of our COph100 dataset files.}
    \label{fig:foldersample}
\end{figure}

\section*{Data Records}
The files available on Figshare \cite{hu2024Coph100}(\href{https://doi.org/10.6084/m9.figshare.27061084}{https://doi.org/10.6084/m9.figshare.27061084}) and Github (\href{https://gaiakoen.github.io/yanhu/research/Retinal\_Image\_Registration}{https://gaiakoen.github.io/yanhu/rese\\arch/Retinal\_Image\_Registration}) include image folders, a data summary in xlsx format \textcolor{black}{ and a Python script for extracting corresponding images from the original dataset.} Our COph100 dataset stands apart from the original classification dataset\cite{timkovivc2024retinal} in three significant ways: Firstly, it is tailored for image registration to monitor disease progression, rather than for classification. Secondly, we have carefully selected a subset of images (325 out of 6,004) from the original dataset, following strict inclusion-exclusion criteria. Finally, we provide 10 point pairs for each image pair, complete with segmented vessel data, making this the largest publicly available retinal image registration dataset for infants to date.

The published dataset comprises 100 eyes, with a total of 491 pairs of images. The details of the examinations are listed in Table \ref{tab:examinationlist}. \textcolor{black}{There are 33 eyes containing 2 examinations, 43 eyes containing 3 examinations, 14 eyes containing 4 examinations, and so on. For the folder name}, the folders labeled '003' and '003-1' indicate that we selected two eyes from the same patient. The folder labeled '04' suggests that one of the patient's eyes was not suitable for registration according to our inclusion and exclusion criteria. 
The structure of the files within our COph100 dataset is illustrated in Figure \ref{fig:foldersample}. Taking the '005' folder as an example, each patient undergoes three examinations, resulting in three distinct images. For each image, the corresponding data includes point pairs (stored in .json files), segmented vessel data (stored in .nii.gz files), vessel masks (stored in .png files), and vessel overlay images (also stored in .png files). 

\begin{figure}[!h]
    \centering
    \includegraphics[width=0.8\textwidth]{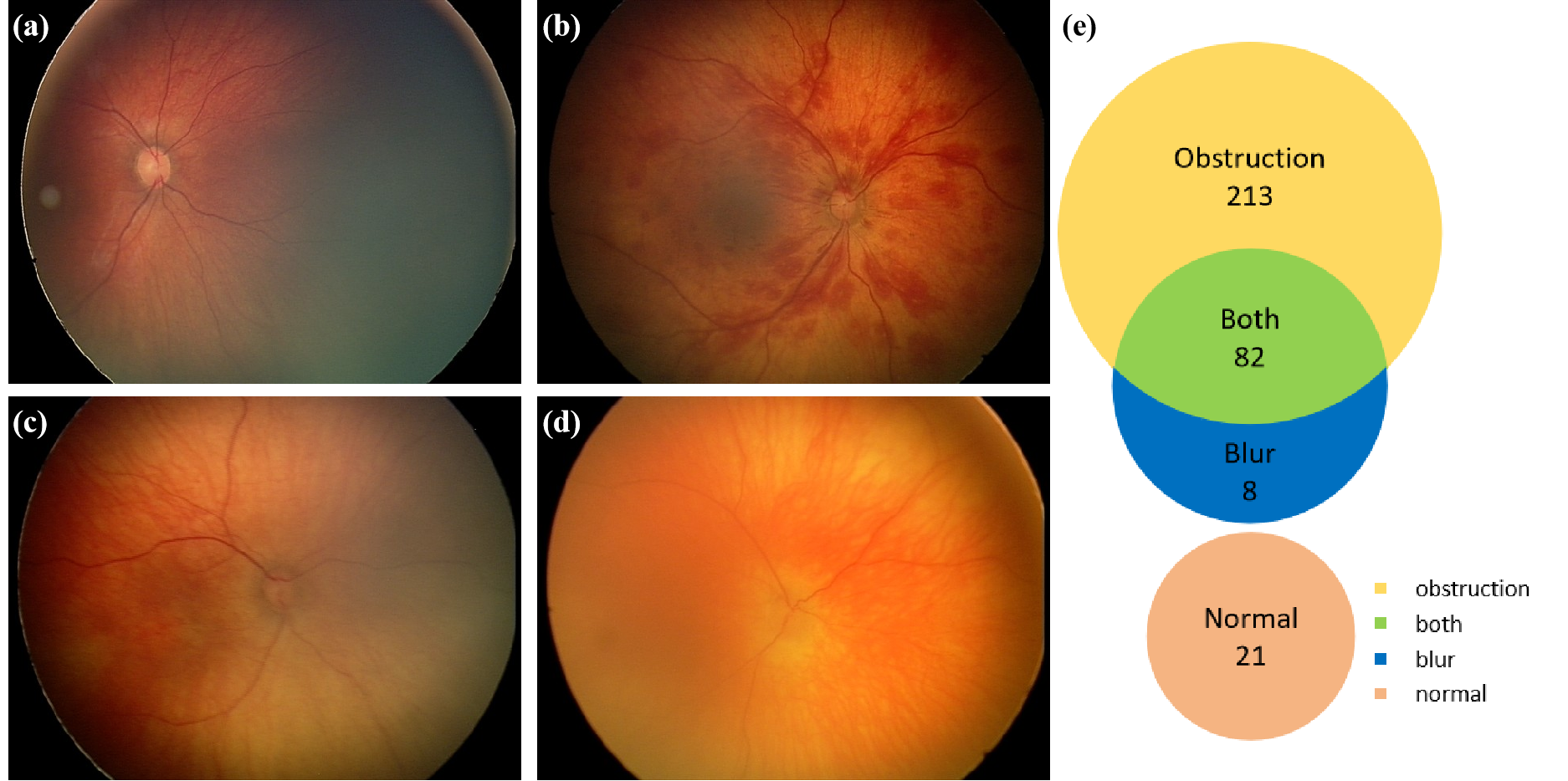}
    \caption{Statistics and examples of the challenge involving blur and obstruction.}
    \label{fig:imageq_bo}
\end{figure}

\begin{figure}
    \centering
    \includegraphics[width=0.8\textwidth]{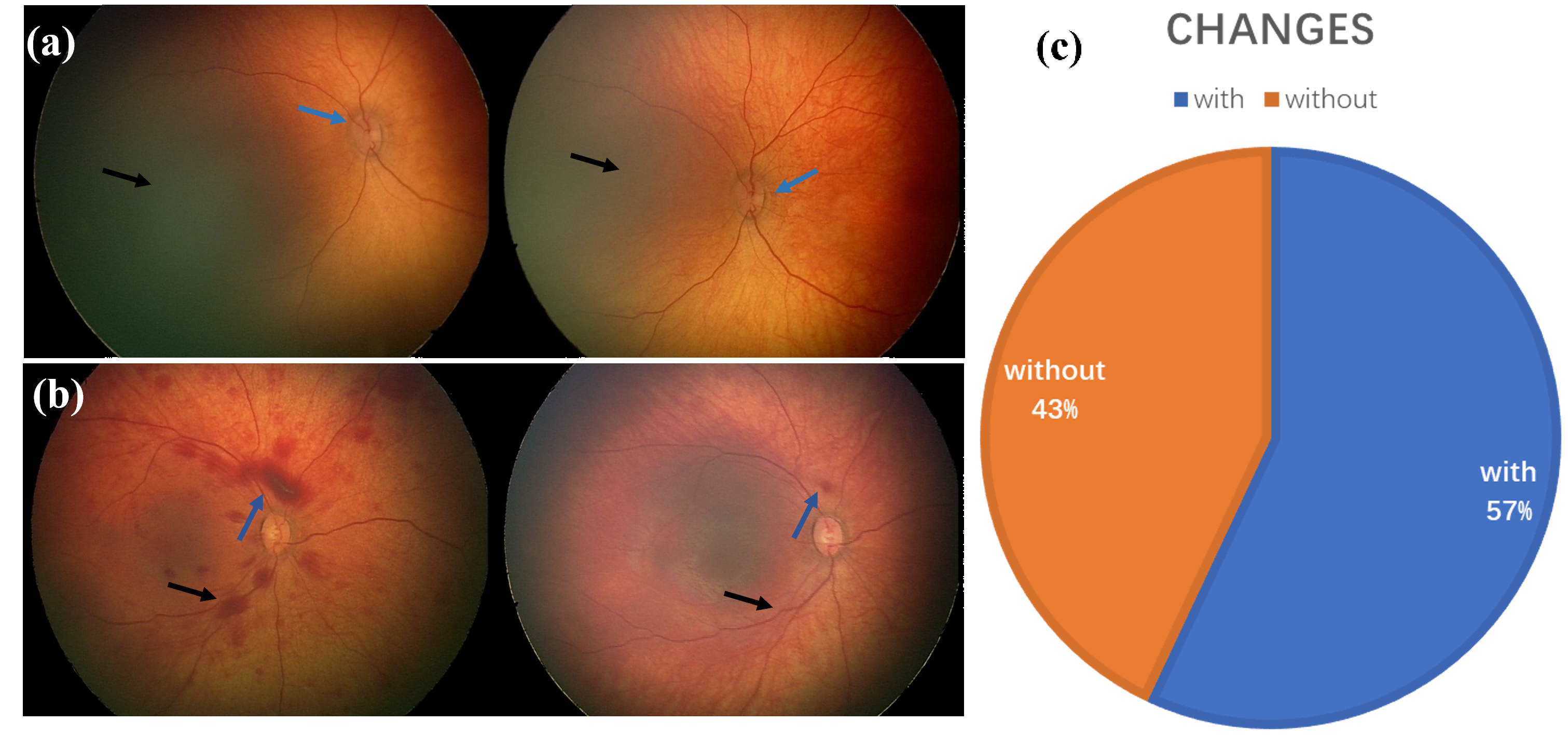}
    \caption{Statistics and examples of the challenge involving changes.}
    \label{fig:imageq_change}
\end{figure}

\section*{Technical Validation}
\subsubsection*{Image quality evaluation}
In order to compound the complexity of image registration tasks, certain artifacts are present in fundus images, which can significantly impact the training phase of registration algorithms. \textcolor{black}{Research on color fundus image quality assessment exists\cite{fu2019evaluation,liu2023deepfundus,shen2020domain}. However, these algorithms cannot be applied to our dataset due to unavailable code, privacy constraints, or differences between their datasets and our infant images.} Consequently, our research team, in collaboration with professional ophthalmologists, has undertaken the evaluation of image quality. This study focuses on three primary challenges: (1) Blur, which is induced by eye movement or out-of-focus conditions; (2) Obstruction, caused by occlusion from disease-related lesions, hemorrhages, patient non-cooperation, probe rotation issues during capture, or turbid refractive media; and (3) Changes, encompassing disease progression, anatomical variations in the visual field, shifts in illumination and color, as well as spatial overlap.

In our dataset of 100 eyes with 325 images, we assessed the quality of each image with respect to blur and obstruction, as well as variations within each eye. Many images presented with multiple challenges. Fig.\ref{fig:imageq_bo} provides statistics and examples illustrating the issues of blur and obstruction present in our dataset. Out of the 325 images, 213 images were affected by obstruction, and 100 images suffered from blur. A total of 82 images experienced both challenges simultaneously, with only 21 images being relatively clear. Consequently, we believe that overcoming the low accuracy of registration caused by obstruction and blur will be an important research challenge for the future. Fig.\ref{fig:imageq_bo}(a) and (b) depict images obstructed by turbid refractive media and hemorrhages, respectively. Fig.\ref{fig:imageq_bo}(c) represents images affected by both blur and obstruction, while Fig.\ref{fig:imageq_bo}(d) illustrates the challenge of blur alone.

Fig.\ref{fig:imageq_change} presents the statistics and examples of the challenges related to changes observed in our COph100 dataset. Over half of the eyes exhibit various changes. In Fig.\ref{fig:imageq_change}(a), the areas of obstruction differ between the two images, and there are notable rotational changes between them. In Fig.\ref{fig:imageq_change}(b), the regions marked by arrows show differences. Prominent hemorrhages are present in the left image during the first examination, but they have almost resolved in the right image by the third examination.

\subsubsection*{Registration evaluation}
To assess the registration performance of existing methods on our dataset, we follow the approach outlined in paper\cite{truong2019glampoints} and report on three categories of results: the ratios of failed, inaccurate and acceptable registrations. For each pair of images in our dataset, the early examination image $I_q$ is used as the query image, while the late examination image acts as the reference $I_r$. A registration attempt is classified as failed if the number of feature matches is less than 4. For successful registrations that obtain a homography, the quality is evaluated using two metrics: the Median Error (MEE) and the Maximum Error (MAE) of the distances between the matched points $M_q$ in the query image and their corresponding estimated positions $E_r$ in the reference image. Registrations are deemed acceptable if both MEE is less than 20 and MAE is less than 50; otherwise, they are labeled as inaccurate. In addition to these measures, we report the Root Square Error (RMSE) for corresponding points after transformation within the acceptable category. Furthermore, we include the Area Under the Curve (AUC) metric \cite{hernandez2017fire} to provide a comprehensive reflection of each method's overall performance.

To comprehensively assess the performance of existing algorithms on our dataset, we conducted experiments based on traditional machine learning and deep learning based algorithms. Traditional algorithms include SIFT\cite{lowe2004distinctive}, GDB-ICP\cite{yang2007registration} \textcolor{black}{(\href{https://vision.cs.rpi.edu/gdbicp/exec/}{https://vision.cs.rpi.edu/gdbicp/exec/})}, REMPE\cite{hernandez2020rempe}\textcolor{black}{(\href{https://projects.ics.forth.gr/cvrl/rempe/}{https://projects.ics.forth.gr/cvrl/rempe/})}. Deep learning-based algorithms include a multitude of algorithms trained on natural scene images (Superpoint\cite{detone2018superpoint}, GLAMpoints\cite{truong2019glampoints}, R2D2\cite{revaud2019r2d2}, SuperGlue\cite{sarlin2020superglue}, LoFTR\cite{sun2021loftr}) and specifically trained on fundus images (SuperRetina\cite{liu2022semi}, Swin U-SuperRetina\cite{nasser2024reverse}, LK-SuperRetina\cite{nasser2024reverse}, and SuperJunction\cite{wang2024superjunction}). \textcolor{black}{We adopted the same testing method as the FIRE dataset, which involves using all the pre-trained models from the official platforms to test on the COph100 dataset.}

Moreover, in light of the recent focus on vascular information in image registration studies, we have adopted a similar approach in our experiments. We utilize the segmentation results of image pairs as the query and reference images to specifically assess the performance of existing registration methods on vascular segmentation images. This choice allows us to evaluate the methods' effectiveness in contexts where vascular alignment is crucial. Note that all our evaluations are conducted at the original image resolution of $640\times 480$ pixels.

\begin{table}[!h]
    \centering
    \caption{Experiment results without (with) segmentation mask based on existing traditional machine learning and deep learning based methods.}
    \label{tab:combined_baseline}
    \begin{tabular}{llcccc} 
        \toprule
                                     & Method                 & Failed [\%]            & Inaccurate [\%]           & RMSE $\downarrow$         & mAUC $\uparrow$        \\
        \midrule
        \multirow{3}{*}{Traditional} & SIFT                   & 0.61 (\textbf{0})       & 86.35 (59.47)             & 9.585 (6.137)             & 0.092 (0.364)          \\
                                     & GDB-ICP                & \textbf{0} (1.22)       & 22.20 (\textbf{2.24})     & 6.096 (\textbf{4.219})    & 0.616 (\textbf{0.834})  \\
                                     & REMPE                  & \textbf{0} (100)         & 30.14 (--)                & 6.094 (--)                & 0.559 (--)              \\
        \midrule
        \multirow{8}{*}{Deep Learning} & SuperPoint           & \textbf{0} (\textbf{0}) & 10.79 (6.72)              & 5.075 (4.604)             & 0.758 (0.807)           \\
                                     & GLAMpoints             & \textbf{0} (\textbf{0}) & 28.51 (25.66)             & 6.540 (5.622)             & 0.564 (0.618)           \\
                                     & R2D2                   & \textbf{0} (\textbf{0}) & 35.03 (3.05)              & 5.546 (5.359)             & 0.529 (0.721)           \\
                                     & SuperGlue              & 0.20 (6.72)             & 4.28 (6.72)               & 5.079 (4.880)             & \textbf{0.809} (0.727)           \\
                                     & LoFTR                  & \textbf{0} (\textbf{0}) & 4.89 (6.11)               & 5.853 (4.531)             & 0.757 (0.807)           \\
                                     & SuperRetina            & 36.66 (51.73)           & \textbf{2.85} (36.25)     & 4.995 (7.761)             & 0.506 (0.093)           \\
                                     & Swin U-SuperRetina     & 3.67 (\textbf{0})       & 16.29 (28.11)             & 5.428 (6.009)             & 0.655 (0.562)           \\
                                     & LK-SuperRetina         & 40.53 (42.36)           & 4.28 (26.07)              & \textbf{4.841} (6.870)    & 0.462 (0.256)           \\
                                     & SuperJunction          & \textbf{0} (100)        & 10.79 (--)                & 4.964 (--)       & 0.755 (--)      \\
        \bottomrule
    \end{tabular}
\end{table}

Table \ref{tab:combined_baseline} summarizes the test results, where the values inside and outside the parentheses represent the results with and without segmentation information, respectively. For traditional methods, GDB-ICP demonstrates strong performance with the lowest failure rate (0\%) and an inaccuracy rate of 22.20\% when no segmentation is applied. It further enhances its performance on vascular segmentation images, achieving a low failure rate of 1.22\% and decreasing the inaccuracy rate to 2.24\%. When segmentation is used, its performance even surpasses that of the deep learning-based methods. SIFT shows a moderate improvement with segmentation, reducing both the failure rate (from 0.61\% to 0\%) and the inaccuracy rate (from 86.35\% to 59.47\%). In contrast, the REMPE method, which depends on 3D modeling of the fundus for registration, fails completely when segmentation information is used, as the segmentation data does not support the necessary 3D modeling process.

\begin{figure}
    \centering
    \includegraphics[width=0.9\textwidth]{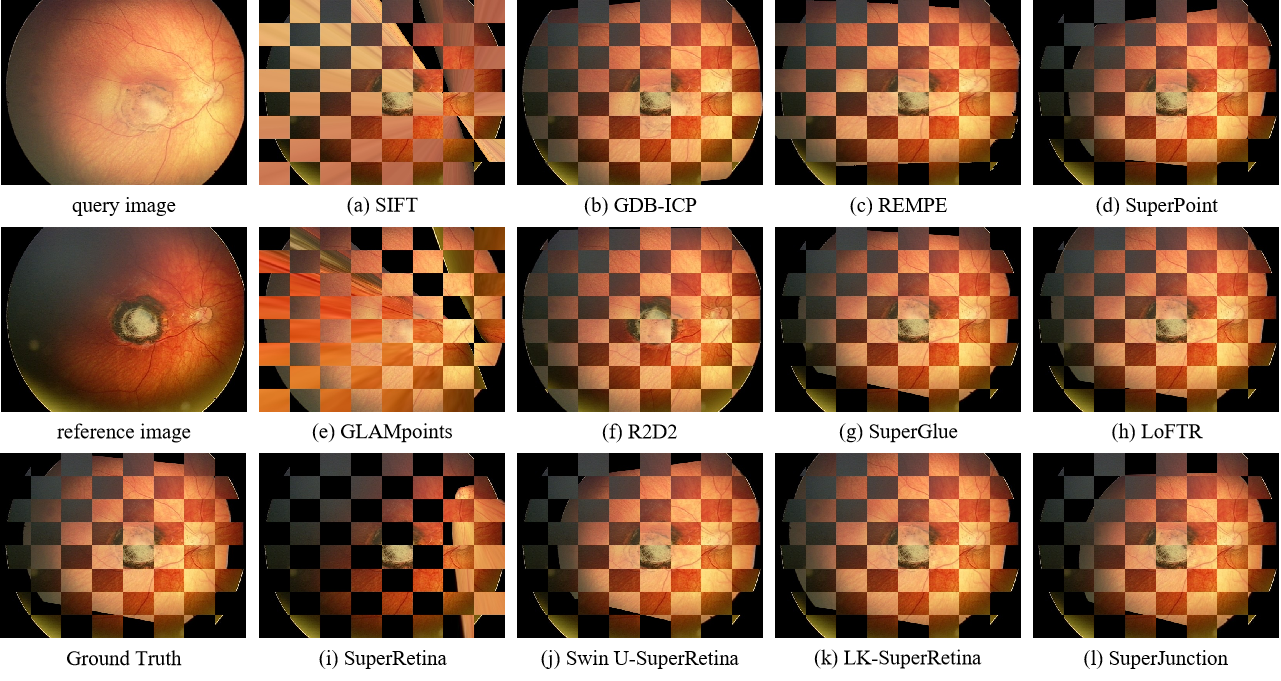}
    \caption{Mosaic visualization of registration results. (a) SIFT (b) GDB-ICP (c) REMPE (d) SuperPoint (e) GLAMpoints (f) R2D2 (g) SuperGlue (h) LoFTR (i) SuperRetina (j) Swin U-SuperRetina (k) LK-SuperRetina (l) SuperJunction}
    \label{fig:visualization}
\end{figure}

Deep learning methods such as SuperPoint, SuperGlue, LoFTR, and SuperJunction have delivered impressive results across all metrics. In contrast, methods including SuperRetina, Swin U-SuperRetina, and LK-SuperRetina have shown higher failure rates and inaccuracies, leading to poor overall performance. When segmentation results are incorporated, the performance of these models further declined, suggesting that models trained exclusively on the fundus dataset may lack robustness when applied to different datasets. Generally, methods trained in natural image environments have demonstrated improved accuracy with the addition of segmentation, except for SuperGlue. This discrepancy might be attributed to the fact that segmentation results can reduce the distinctiveness of descriptors in SuperGlue, which hinders effective matching. We noted that segmentation information tends to improve the registration accuracy for most methods, which underscores the potential benefits of exploring new strategies that more effectively integrate vascular segmentation data to enhance the performance of fundus image registration tasks. In Fig.\ref{fig:visualization}, we present the registration results using a checkerboard mosaic visualization. It is evident that SIFT, GLAMpoints, and SuperRetina clearly failed. \textcolor{black}{Even the other methods, despite performing better, still exhibit noticeable misalignments in the vascular structures compared to the ground-truth registration results.} This highlights the need for registration methods to further improve accuracy and achieve complete alignment of the blood vessels in the images for more reliable outcomes.
Therefore, the results from the table indicate that our proposed COph100 dataset represents certain challenges for both traditional machine learning and deep learning methods. There is considerable potential for enhancement in the future.

\section*{Usage Notes}
The COph100 dataset is intended to serve as a resource for researchers working in the field of medical image analysis, particularly those interested in image registration, disease progression analysis. The COph100 dataset presented in this paper can be downloaded through the link mentioned above. 

\section*{Code availability}
The code associated with this study are all publicly available. We do not develop any new code. \\ 
SIFT detector plus RootSIFT descriptor, using OpenCV APIs. \\ 
GDB-ICP, obtained from \href{https://vision.cs.rpi.edu/gdbicp/exec/}{https://vision.cs.rpi.edu/gdbicp/exec/}. \\ 
REMPE, obtained from \href{https://projects.ics.forth.gr/cvrl/rempe/}{https://projects.ics.forth.gr/cvrl/rempe/}. \\ 
SuperPoint, trained on MS-COCO, obtained from \href{https://github.com/rpautrat/SuperPoint}{https://github.com/rpautrat/SuperPoint}. \\ 
GLAMpoints, obtained from \href{https://github.com/PruneTruong/GLAMpoints_pytorch}{https://github.com/PruneTruong/GLAMpoints\_pytorch}.\\ 
R2D2, obtained from \href{https://github.com/naver/r2d2}{https://github.com/naver/r2d2}. \\ 
SuperGlue, trained on ScanNet, obtained from \href{https://github.com/magicleap/SuperGluePretrainedNetwork}{https://github.com/magicleap/SuperGluePretrainedNetwork}. \\ 
LoFTR, obtained from \href{https://github.com/zju3dv/LoFTR}{https://github.com/zju3dv/LoFTR}. \\ 
SuperRetina, obtained from \href{https://github.com/ruc-aimc-lab/superretina}{https://github.com/ruc-aimc-lab/superretina}. \\ 
Swin UNet and LK SuperRetina, obtained from \href{https://github.com/NiharGupte/ReverseKnowledgeDistillation}{https://github.com/NiharGupte/ReverseKnowledgeDistillation}. \\ 
SuperJunction, obtained from \href{https://github.com/AdamWang0224/SuperJunction}{https://github.com/AdamWang0224/SuperJunction}. \\ 
SS-MAF, obtained from \href{https://github.com/Qsingle/imed\_vision}{https://github.com/Qsingle/imed\_vision}.

\bibliography{sample}

\section*{Acknowledgements}
This work was supported in part by The National Natural Science Foundation of China (82102189, 82272086 and 62401246), Shenzhen Stable Support Plan Program (20220815111736001), and Special Funds for the Cultivation of Guangdong College Students’Scientific and Technological Innovation (“Climbing Program”Special Funds) (pdjh2024b331).

\section*{Author contributions statement}
Y.H. data collection and labelling, idea maker, image quality check and draft preparation; M.G. data labelling and experiments; Z.Q. image segmentation;J.L., H.S.,X.Z. and H.L. data labelling, GT and image quality check; M.Y. and H.L. medical consultation, segmentation results and image quality check; J.L. supervision. All authors reviewed the manuscript and agreed to the submitted version of the manuscript.
\section*{Competing interests}
The authors declare no competing interests.

\end{document}